\documentclass{isprs} 
\usepackage{subfigure}
\usepackage{setspace}
\usepackage{geometry} 
\usepackage{epstopdf}
\usepackage[labelsep=period]{caption}  
\usepackage[british]{babel} 
\usepackage[hang]{footmisc}

\usepackage{url}
\usepackage{amsmath}
\usepackage{booktabs}
\usepackage{graphicx}
\usepackage{float}
\usepackage{amssymb}

\usepackage{enumitem}

\begin{document}

\title{MambaPanoptic: A Vision Mamba-based Structured State Space Framework for Panoptic Segmentation}
\date{}

\author{
 Qing Cheng\textsuperscript{1,2}\footnotemark[1]   \ , 
 Damiano Bertolini\textsuperscript{1,3}\thanks{Equal Contribution} \ , 
 Wei Zhang\textsuperscript{4}, 
 Dong Wang\textsuperscript{5}, 
 Niclas Zeller\textsuperscript{6}, 
 Daniel Cremers\textsuperscript{1,2}
}

\address{
	\textsuperscript{1 }Technical University of Munich, (qing.cheng, cremers)@tum.de\\
	\textsuperscript{2 }Munich Center for Machine Learning (MCML)\\
	\textsuperscript{3 }Polytechnic University of Milan, damiano.bertolini@mail.polimi.it\\
	\textsuperscript{4 }University of Stuttgart, wei.zhang@ifp.uni-stuttgart.de\\
    \textsuperscript{5 }Wuhan University, timdong@whu.edu.cn\\
	\textsuperscript{6 }Karlsruhe University of Applied Sciences, niclas.zeller@h-ka.de\\
}

\abstract{

Panoptic segmentation requires the simultaneous recognition of countable \textit{thing} instances and amorphous \textit{stuff} regions, placing joint demands on long-range context modelling, multi-scale feature representation, and efficient dense prediction. Existing convolutional and transformer-based methods struggle to satisfy all three requirements concurrently: convolutional architectures are limited in their capacity to model long-range dependencies, while transformer-based methods incur quadratic computational cost that is prohibitive at high resolutions. In this paper, we propose MambaPanoptic, a fully Mamba-based panoptic segmentation framework that addresses these limitations through two principal contributions. First, we introduce MambaFPN, a top-down feature pyramid that leverages Mamba blocks to generate globally coherent, multi-scale feature representations with linear computational complexity. Second, we adopt a PanopticFCN-style kernel generator that produces unified \textit{thing} and \textit{stuff} kernels for proposal-free panoptic prediction, enhanced by a QuadMamba-based feature refinement module applied at multiple network stages. Experiments on the Cityscapes and COCO panoptic segmentation benchmarks demonstrate that MambaPanoptic consistently outperforms PanopticDeepLab and PanopticFCN under comparable model sizes, and matches or surpasses Mask2Former on Cityscapes in PQ and AP while requiring fewer parameters.

}

\keywords{Panoptic Segmentation, Structured State Space Models, Vision Mamba}

\maketitle

\section{INTRODUCTION}\label{INTRODUCTION}

Scene understanding is one of the most important computer vision topics, and panoptic segmentation is one of the most complete and fine-grained tasks in this domain. Panoptic segmentation unifies semantic segmentation and instance segmentation: it estimates a class label for each pixel, and detects the instances and assigns the instance label to them \cite{Kirillov2019Panoptic}. This rich and complete scene understanding is so beneficial to various real-world vision-based applications: e.g. autonomous driving, mobile robotics, AR/VR systems, etc, that it has attracted significant attention \cite{Kirillov2019PanopticFPN,Cheng2020PanopticDeepLab,Li2021PanopticFCN,Wang2021MaXDeepLab,Li2022PanopticSegFormer,Cheng2022Mask2Former}.

Panoptic segmentation remains challenging because it must simultaneously recognize object instances and delineate amorphous background regions at the pixel level. This requires feature representations that preserve fine local details for accurate boundaries and small objects, while also capturing long-range context to reason about large stuff regions and complex scene layouts. In practice, these competing requirements make panoptic segmentation particularly demanding for high-resolution real-world images.

Conventional methods for panoptic segmentation have depended mostly on Convolutional Neural Networks (CNNs) \cite{Kirillov2019PanopticFPN} or, more recently, vision transformers \cite{carion2020end}. Convolutional neural networks are effective at capturing local structures through their translation-equivariant inductive bias, but their receptive fields are inherently bounded, limiting their ability to model long-range spatial dependencies. Transformers overcome this limitation through global self-attention, yet do so at a quadratic computational and memory cost with respect to sequence length, which is a significant bottleneck for high-resolution inputs and resource-constrained deployments. Furthermore, the absence of strong inductive biases in transformers typically necessitates large-scale training data and long training time compared to comparably-sized convolutional models.

Structured State Space Models (SSMs), and in particular Mamba architectures, have recently emerged as a promising alternative. Mamba models long-range dependencies through selective state-space dynamics while maintaining linear complexity with respect to the input length \cite{Gu2023Mamba}. Building on this idea, Vision Mamba extends state-space modeling to visual data by processing image patches with selective bidirectional scanning, enabling both local and global context aggregation. Variants such as the 2D Selective Scan further improve spatial modeling by capturing dependencies along different spatial directions while preserving neighborhood structure \cite{Zhu2024VisionMamba}.

In this paper, we propose MambaPanoptic, a fully Mamba-based architecture for panoptic segmentation. Our model integrates a hierarchical Vision Mamba encoder with a novel Mamba Feature Pyramid Network (MambaFPN) — a top-down multi-scale feature pyramid that employs Mamba blocks to produce globally coherent, spatially rich feature representations. Building on this encoder, we adopt a PanopticFCN-style kernel generator \cite{Li2021PanopticFCN} that produces unified \textit{thing} and \textit{stuff} kernels for proposal-free panoptic prediction, and further introduce a QuadMamba-based \cite{Xie2024QuadMamba} feature refinement module to progressively improve feature quality at multiple network stages. To the best of our knowledge, MambaPanoptic is the first end-to-end Mamba-based architecture designed specifically for panoptic segmentation. SegMAN \cite{Fu2025SegMAN} only replaces the ResNet50 \cite{he2016deep} with its encoder as the backbone of Mask DINO \cite{li2023mask} for panoptic segmentation. Extensive experiments on the Cityscapes \cite{Cordts2016CityscapesCVPR} and COCO \cite{Lin2014COCO} benchmarks demonstrate that MambaPanoptic consistently outperforms CNN-based base\-lines, including Panoptic-DeepLab \cite{Cheng2020PanopticDeepLab} and PanopticFCN \cite{Li2021PanopticFCN}, and matches or surpasses the transformer-based Mask2Former \cite{Cheng2022Mask2Former} on Cityscapes in PQ and AP, while requiring fewer parameters.

In sum, our contributions are as follows: 
\begin{itemize}
    \item We present MambaPanoptic, the first fully Mamba-based end-to-end panoptic segmentation architecture with strong performance;
    \item We propose MambaFPN, a linear-time multi-scale feature pyramid module that jointly provides global context and fine spatial detail for panoptic prediction;
    \item We also introduce the QuadMamba-based feature refinement module (MFRM) that efficiently enhances feature representations across multiple stages of the network;
    \item We conduct comprehensive experiments and ablation studies of the proposed method on Cityscapes and COCO panoptic benchmarks, demonstrating competitive performance against representative CNN-based and transformer-based baselines.
\end{itemize}

\section{RELATED WORK}\label{RELATED_WORK}

\subsection{CNN-based Panoptic Segmentation}

Early panoptic segmentation methods were largely built upon CNN-based instance segmentation frameworks, typically by combining an instance branch with a semantic segmentation branch. A representative example is Panoptic FPN \cite{Kirillov2019PanopticFPN}, which extends Mask R-CNN \cite{he2017mask} with an additional fully convolutional branch for \textit{stuff} prediction and merges the two outputs through heuristic post-processing. UPSNet \cite{Xiong2019UPSNet} further improves this paradigm by adopting a shared deformable-convolution backbone and a lightweight panoptic head with an additional unknown class to better handle conflicts between instance and semantic predictions. Other dual-branch architectures, such as TASCNet \cite{Li2018TASCNet} and AUNet \cite{Li2018AUNet}, also model \textit{thing} and \textit{stuff} categories separately, but still rely on explicit fusion procedures that may limit the coherence of the final panoptic output.

Another line of work explores bottom-up, proposal-free formulations. Panoptic-DeepLab \cite{Cheng2020PanopticDeepLab} predicts dense semantic maps together with instance-aware cues, including object center heatmaps and per-pixel offset vectors. This design avoids region proposals and clustering over object candidates while preserving strong spatial consistency in the predicted masks. 

More recently, unified single-head CNN architectures have been proposed to further simplify the pipeline. PanopticFCN \cite{Li2021PanopticFCN}, for example, formulates both \textit{thing} instances and \textit{stuff} regions as kernel-based predictions, enabling direct mask generation from dense feature maps without relying on bounding boxes or proposal generation. Similarly, the Category-Instance Embedding approach \cite{Gao2020CIAE} learns per-pixel embeddings that jointly encode semantic and instance information within a unified representation space.

Overall, CNN-based methods remain attractive due to their strong locality bias, efficiency, and relatively simple dense prediction pipelines. However, their limited ability to model long-range dependencies can be a disadvantage for panoptic segmentation, where both large-scale scene context and fine-grained spatial details are crucial.

\subsection{Transformer-based Panoptic Segmentation}

Transformer-based methods have recently become a dominant paradigm for panoptic segmentation by reformulating the task as mask classification. Instead of predicting dense outputs through separate semantic and instance branches, these methods learn a set of queries, each associated with a class label and a segmentation mask. MaskFormer \cite{Cheng2021MaskFormer} is a representative framework in this direction, showing that panoptic segmentation can be unified through a fixed set of mask queries refined with attention-based decoding. This formulation eliminates the need for handcrafted fusion between semantic and instance predictions and provides a clean end-to-end training framework.

Building on this idea, Mask2Former \cite{Cheng2022Mask2Former} introduces masked attention, which restricts each query to attend primarily to its predicted spatial support. Combined with a strong multi-scale pixel decoder based on deformable attention, this design improves both efficiency and accuracy, and has become a strong baseline across semantic, instance, and panoptic segmentation tasks. Other transformer-based approaches pursue similar goals with different architectural choices. MaX-DeepLab \cite{Wang2021MaXDeepLab} integrates a mask transformer with a CNN backbone and jointly learns pixel features and mask embeddings in a unified framework. Panoptic SegFormer \cite{Li2022PanopticSegFormer} follows a related query-based design, while introducing separate queries for \textit{thing} and \textit{stuff} categories as well as deep supervision to stabilize optimization.

Compared with CNN-based models, transformer-based methods offer stronger global reasoning and a more unified prediction framework. However, these benefits typically come at the cost of higher computational and memory complexity, particularly for high-resolution dense prediction. This trade-off motivates the investigation of alternative architectures that can retain strong long-range modeling while remaining computationally efficient.

\subsection{Vision Mamba-based Segmentation}
State-space models (SSMs) led by Mamba \cite{Gu2023Mamba} have emerged as efficient alternatives to attention for long-range context modeling with linear-time scaling via input-dependent selective state updates. The 2D Selective Scan of VMamba \cite{Liu2024VMamba} and the bi-directional SSM of Vision Mamba \cite{Zhu2024VisionMamba} bridge 1D sequence modeling to 2D images and demonstrate strong performance on vision tasks, e.g., image classification, object detection, and semantic segmentation. The follow-ups, e.g., MambaVision \cite{hatamizadeh2025mambavision}, GroupMamba \cite{shaker2025groupmamba}, MobileMamba \cite{he2025mobilemamba}, SegMAN \cite{Fu2025SegMAN}, focus on improving the performance of semantic and instance segmentation.
For aerial image segmentation, Mamba-style state-space models have demonstrated notable progress: e.g., dual-branch RS3Mamba \cite{ma2024rs}, encoder–decoder Samba \cite{ren2024samba}, large-VHR RS-Mamba \cite{zhao2024rs}, and lightweight UNetMamba \cite{zhu2024unetmamba}. Meanwhile, the medical variants, e.g. U-Mamba \cite{ma2024u}, VM-UNet \cite{10.1145/3767748}, and SegMamba \cite{Xin_SegMamba_MICCAI2024}, also validate the effectiveness of Mamba in encoder–decoder segmentation designs, reinforcing its generality for dense prediction tasks.
Collectively, these studies position Vision Mamba backbones as promising for high-resolution semantic segmentation.

Motivated by these advances, we propose a Mamba-based architecture for panoptic segmentation. Our method aims to combine the efficiency and long-range modeling capability of state space models with a unified panoptic prediction framework, providing an alternative to both CNN-based and transformer-based designs.

\begin{figure*}[h]
\centering
\includegraphics[width=1.0\textwidth]
{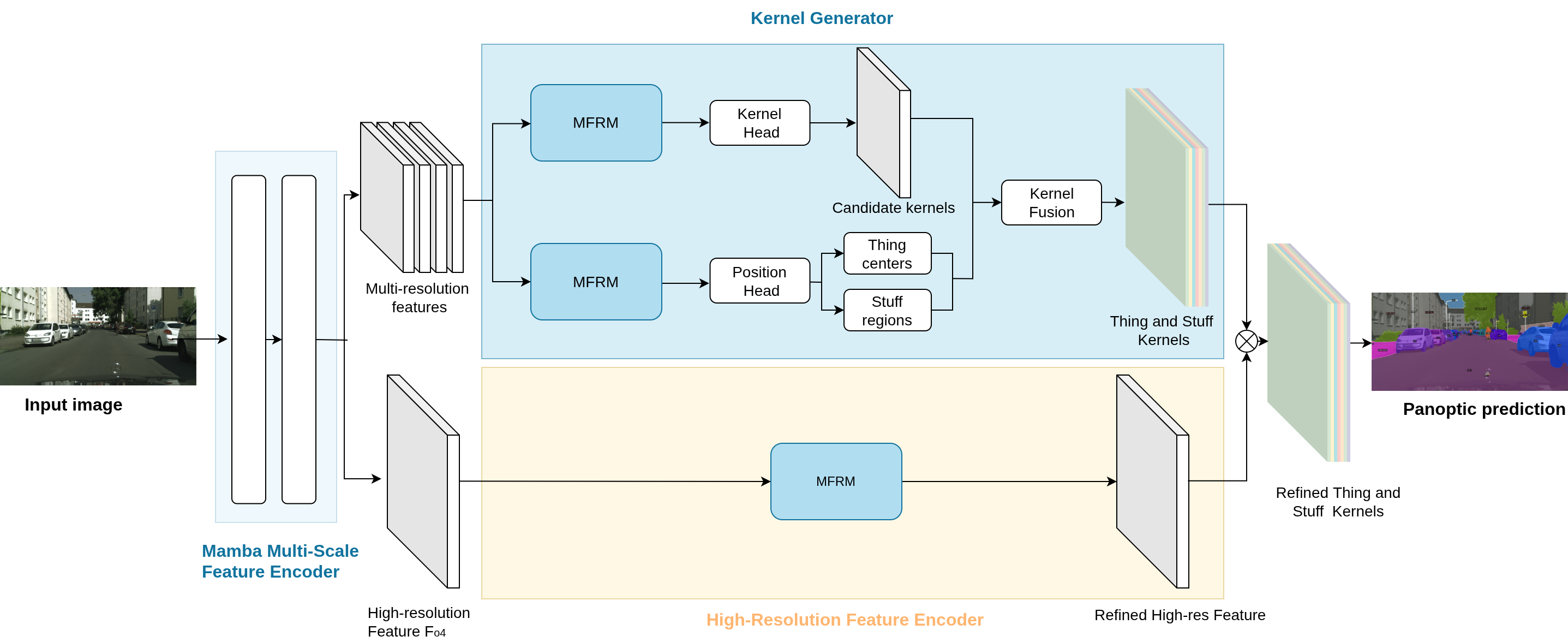}
\caption{The architecture of the proposed Mamba-based panoptic segmentation network. 
The MambaFPN takes an image as input the outputs a set of multi-scale feature maps. The kernel generator processes each features and outputs the kernel masks and kernel features at each resolution. Kernel fusion module fuses the kernels from different resolutions into distinct \textit{thing} and \textit{stuff} kernels. The high-resolution feature is refined by the high-resolution feature encoder and then is injected to the kernel features. Finally the panoptic segmentation is estimated based on these final kernels.}
\label{fig:architecture}
\end{figure*}

\begin{figure*}[htbp]
\centering
\includegraphics[width=1.0\textwidth]
{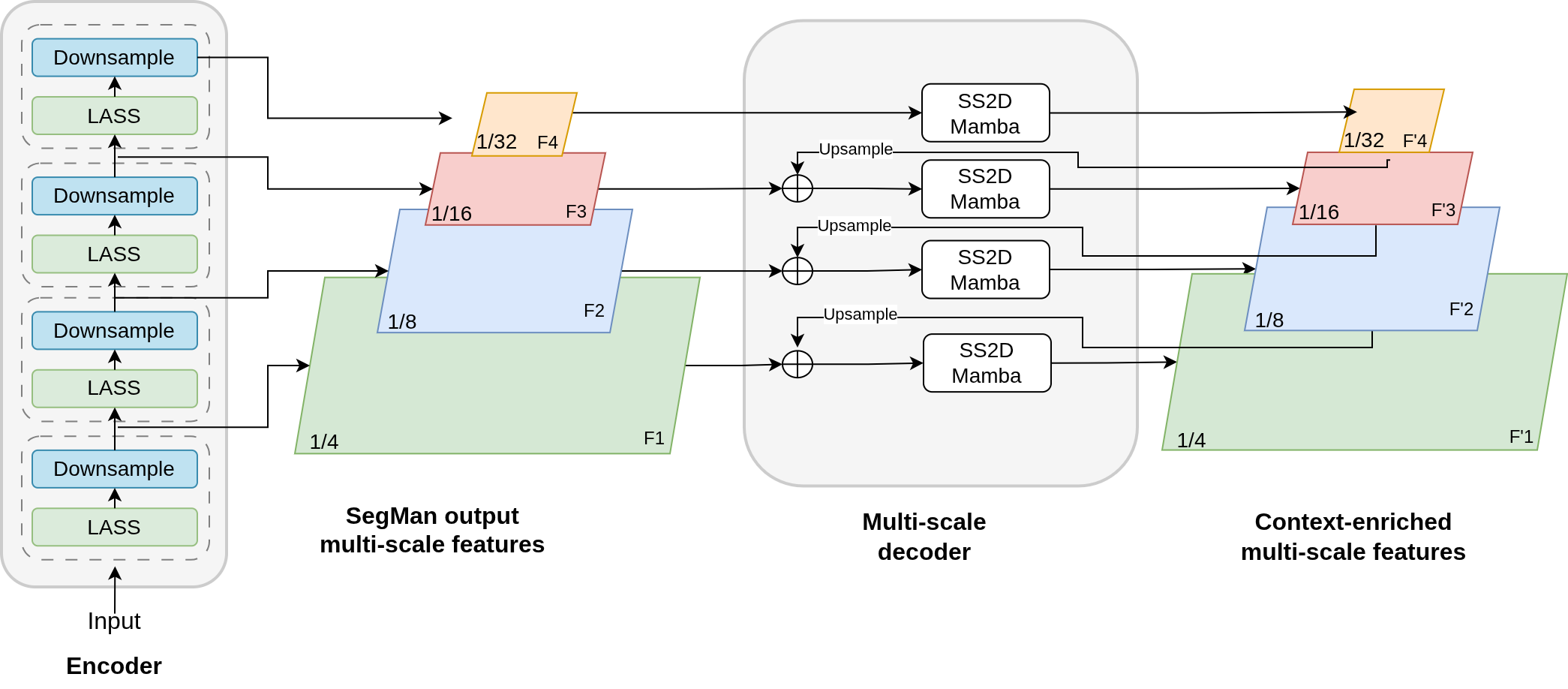}
\caption{The architecture of the proposed Mamba-based multi-scale feature encoder. The SegMan encoder processes the input image and multi-scale features are extracted. These features are further upsampled to fuse the higher feature map and then processed by the SS2D blocks to generate the context-enriched multi-scale features.}
\label{fig:mambafpn}
\end{figure*}

\section{METHOD}\label{METHOD}

In this section, we describe the proposed Mamba-based architecture for panoptic segmentation.

Panoptic segmentation assigns each pixel $p$ a semantic label $c$, $c \in C$, and a unique instance identity $i$ for each object of thing categories. The label space $C$ is divided into thing classes, which correspond to countable object instances, and stuff classes, which correspond to amorphous background regions.

As illustrated in Figure \ref{fig:architecture}, the proposed method is a unified, proposal-free framework for joint \textit{thing} and \textit{stuff} segmentation, comprising two principal components: a Mamba-based multi-scale feature encoder and a panoptic segmentation head. A Mamba feature refinement module is additionally introduced at multiple stages to improve feature representability. Each component is described in detail below.

\subsection{Mamba Multi-Scale Feature Encoder}
\label{encoder}
Panoptic segmentation requires feature representations that preserve fine local structures while also modeling long-range contextual dependencies. To overcome the locality bias of CNNs and the quadratic computation requirement of transformers, we devise a Mamba-based image encoder that aggregates global context with linear complexity with respect to sequence length \cite{Liu2024VMamba}. The proposed Mamba-based multi-scale feature encoder consists of a hierarchical backbone and a top-down feature pyramid, as shown in Figure \ref{fig:mambafpn}.

\subsubsection{Mamba backbone}
We use the SegMAN encoder \cite{Fu2025SegMAN} as the backbone, which is a hierarchical hybrid network designed to jointly capture global context and fine-grained spatial detail with linear computational complexity. The SegMAN encoder consists of four blocks. Each block starts with a strided $3\times3$ convolution for spatial downsampling, followed by  a series of Local Attention and State Space (LASS) blocks for feature processing. Each LASS block is structured as a sequential combination of Neighbourhood Attention and the 2D Selective Scan (SS2D) mechanism \cite{Liu2024VMamba}. Neighbourhood Attention operates within a fixed local window, preserving spatial precision and translational invariance, while SS2D models long-range dependencies by scanning the flattened feature sequence in four orthogonal directions. The outputs of the two sub-modules are combined via a residual connection and a $1\times1$ convolution, enabling effective cross-scale feature interaction.

 \subsubsection{Mamba Feature Pyramid Network}
Panoptic segmentation consists of semantic segmentation for the \textit{stuff} classes and instance segmentation for the object classes. Stuff classes tend to be dense and cover the full image, while object segments tend to be local and varying-scale. These two sub-tasks impose different requirements for the feature property. Thus, the features are required to be globally coherent to cover the large segments, sufficiently high resolution to encode fine structures, efficiently capturing multi-scale information to handle the varying-size objects, and sufficiently rich semantics to enable reliable class prediction \cite{Kirillov2019Panoptic,Kirillov2019PanopticFPN,Li2021PanopticFCN}. To fulfill these requirements, we design a Mamba-based feature pyramid network (MambaFPN) to generate multi-scale image features for the panoptic head, inspired by FPN \cite{lin2017feature}. 

The proposed MambaFPN consists of 4 blocks, each of which fuses the features from two different stages and applies local attention with an SS2D block. The four consecutive blocks of the SegMAN encoder generate four feature maps $\{F_1,F_2,F_3,F_4\}$ at resolutions 1/4, 1/8, 1/16, and 1/32 of the input image, respectively. Starting from the deepest feature map $F_4$, an SS2D block is applied to produce $F'_4$. Each subsequent MambaFPN block then upsamples the feature map $F_i$ and integrates the $F'_{i+1}$ by $1\times1$ convolution and  channel-wise addition, and then refines the fused feature map by an SS2D block. This process yields four context-enriched feature maps $\{F'_1,F'_2,F'_3,F'_4\}$ at scales of 1/4, 1/8, 1/16, and 1/32, which are subsequently passed to the panoptic head for panoptic segmentation. The MambaFPN is visualised in Figure \ref{fig:mambafpn}.

\subsection{Panoptic Head}
\label{head}
The recent advancements in panoptic segmentation favour the kernel-based \cite{Li2021PanopticFCN}, or query-based \cite{Cheng2021MaskFormer} panoptic head for their effectiveness and simplicity due to their proposal-free and unified single-pass prediction \cite{LI2022103283}. In our work, we adopt the kernel-based panoptic head design to avoid the heavy transformer decoder and self-attention in the query-based panoptic head.

Our panoptic head is inspired by PanopticFCN \cite{Li2021PanopticFCN}, which proposes a unified architecture for managing both \textit{thing} and \textit{stuff} prediction by representing them as kernels. We enhance this design with the multi-scale contextual representations produced by MambaFPN and introduce Mamba-based feature refinement at key stages of the head. The panoptic head consists of two branches: a kernel generator and a high-resolution feature encoder. The outputs of both branches are convolved to produce the final panoptic segmentation.

\subsubsection{Kernel Generator}

The kernel generator utilises instance centres and semantic regions to represent \textit{thing} and \textit{stuff} with kernels, respectively. Each kernel can be viewed as a mask with the same resolution as the input feature map. Thus, \textit{thing} and \textit{stuff} have a unified kernel representation.
The kernel generator has two sub-modules: a position head for kernel localization and categorization, and a kernel head for producing kernel weights. The position head first refines the input MambaFPN features and predicts heatmaps $M^t$ for \textit{thing} centres, where the peak of a heatmap represents the \textit{thing} center, and score maps $M^s$ for \textit{stuff} regions, where the high response positions of a score map indicate a \textit{stuff} region. The heatmaps and score maps contribute to the overall kernel masks $M$. The kernel head is designed to generate the features. It first refines the input MambaFPN features and concatenates the relative coordinates to the refined feature map, and applies a lightweight CNN to generate spatially aware kernel features $F$. These features are then indexed by the kernel masks to aggregate the features for each kernel, resulting in a set of \textit{thing} kernel features $F^t$ and a set of \textit{stuff} kernel features $F^s$.

\subsubsection{Kernel Fusion}
The kernel generator works on the four feature maps from the MambaFPN individually, so there can be multiple kernel estimations at different resolutions for the same target. Therefore, we apply a kernel fusion module to integrate the kernels at the different resolutions. We fuse the \textit{thing} and \textit{stuff} kernels differently due to their representations. For \textit{things}, we group kernels across resolutions according to their kernel features: two kernels are identified as the same instance if their cosine similarity is above a given threshold. For \textit{stuff}, we utilize the predicted semantic class and group kernels sharing the same class prediction. The corresponding kernel features are then aggregated via average pooling over the matched kernels. Finally, this yields a unique set of kernel features ${\hat{K}}$ for each input image, consisting of \textit{thing} kernel features $\hat{K}^t$ and \textit{stuff} kernel features $\hat{K}^s$.

\subsubsection{High-Resolution Feature Encoder}
To better preserve the fine structure and boundaries of the \textit{thing} and \textit{stuff} representation, following PanopticFCN \cite{Li2021PanopticFCN}, we utilize the highest-resolution feature map from our MambaFPN to inject high-resolution context into the kernels. Specifically, we first refine the high-resolution feature map, $F'_1$, with our proposed Mamba-based refinement module and use the fused kernel masks $M$ to retrieve the corresponding features and then convolve with the corresponding kernel features $\hat{K}$ to generate the fused kernel features $K$.

\subsubsection{Mamba feature refinement module}
\label{refine}
To refine the features in both the kernel generator and the high-resolution feature encoder, we introduce a Mamba feature refinement module (MFRM) to improve local spatial structure and global contextual awareness, which is built upon a single QuadMamba block \cite{Xie2024QuadMamba}. 
QuadMamba \cite{Xie2024QuadMamba} uses a learnable quadtree-based scanning strategy that adaptively partitions the feature map into regions of different granularity. A partition predictor estimates the locality score of each token and determines whether the corresponding region should be further subdivided, enabling coarse-to-fine adaptive processing. In addition, an omnidirectional window-shifting mechanism promotes feature interaction across region boundaries. Compared with fixed-window partitioning \cite{Huang2024LocalMamba}, this design is better suited to objects with varying scales and irregular shapes. We use this module as a lightweight refinement block to enhance feature representations before kernel prediction and final mask decoding.

\subsubsection{Panoptic prediction}
The fused kernel representations are used for the final panoptic prediction. For the \textit{thing} objects, the model predicts a binary mask and a class label over the \textit{thing} labels, and for \textit{stuff} kernels, it estimates per-pixel segmentation over \textit{stuff} classes. Finally, the \textit{thing} and \textit{stuff} masks are merged into a single panoptic map using the heuristic fusion procedure of PanopticFPN \cite{Kirillov2019PanopticFPN}.

\subsection{Training}
The panoptic prediction task involves two objectives: kernel mask detection and semantic label classification. Accordingly, the model is trained with two complementary loss terms.
For the kernel score maps, Focal loss \cite{lin2017focal} is used for optimization. As the \textit{thing} objects are represented by their centers, we construct the \textit{thing} ground truth as a continuous heatmap generated by placing 2D Gaussian kernels at the center of each object instance. The \textit{stuff} regions for semantic segmentation are formulated commonly with one-hot encoding and bilinearly interpolated to match the resolution of the prediction. The kernel loss is defined as: 

\vspace{-1em}
\begin{equation}
\begin{aligned}
    L_{k} = \frac{1}{N_{t}} \sum_{i}FL(K_i^{t}, Y_i^{t})  +  \frac{1}{N_{s}} \sum_{i}FL(K_i^{s}, Y_i^{s})
\end{aligned}
\end{equation}
where $N_{t}$ is the number of \textit{thing} object kernels and $N_{s}$ is the number of pixels; $K_i$ is the predicted kernel mask and $Y_i$ is the ground truth mask.

For the mask prediction branch, the final mask probabilities $P_i$ are obtained by taking the dot product between the fused kernel features and the refined high-resolution feature map $F'_1$, followed by a sigmoid activation for \textit{thing} masks and softmax for \textit{stuff} regions. Mask prediction is supervised using Dice loss \cite{milletari2016v} in order to handle the imbalanced label distribution inherent in dense segmentation:

\vspace{-1em}
\begin{equation}
\begin{aligned}
    L_{seg} = \frac{1}{N} \sum_{i}Dice(P_i, Y_i) 
\end{aligned}
\end{equation}

The total training loss is a weighted combination of the two terms:
\vspace{-0.5em}
\begin{equation}
\begin{aligned}
    L_{total} = \lambda_{k} L_{k} + \lambda_{seg} L_{seg}
\end{aligned}
\end{equation}

\begin{figure*}[h]
\centering
\setlength{\tabcolsep}{0pt}
\renewcommand{\arraystretch}{0}

\begin{tabular}{@{}cccc@{}}
\includegraphics[width=0.24\textwidth]{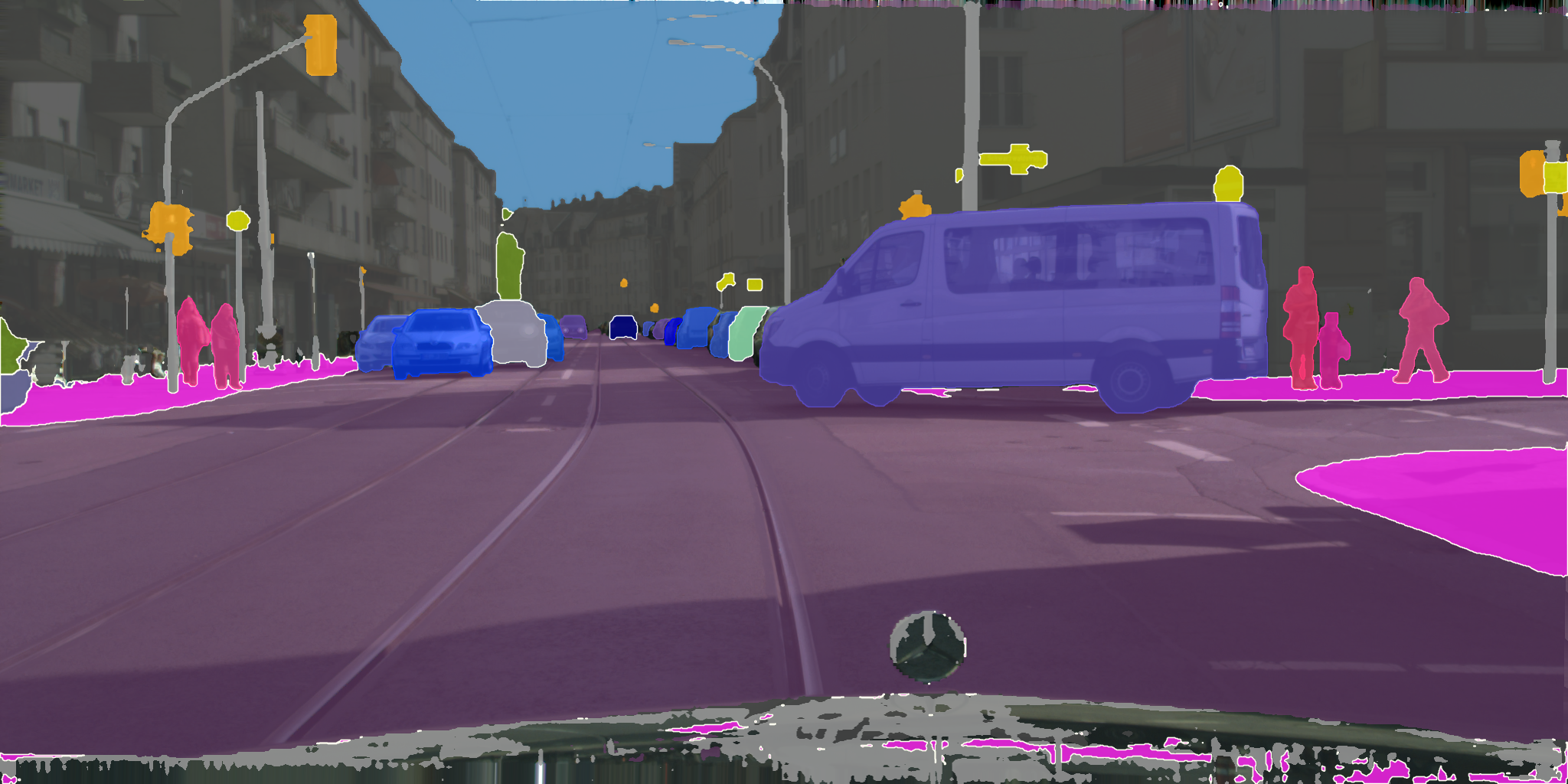} & 
\includegraphics[width=0.24\textwidth]{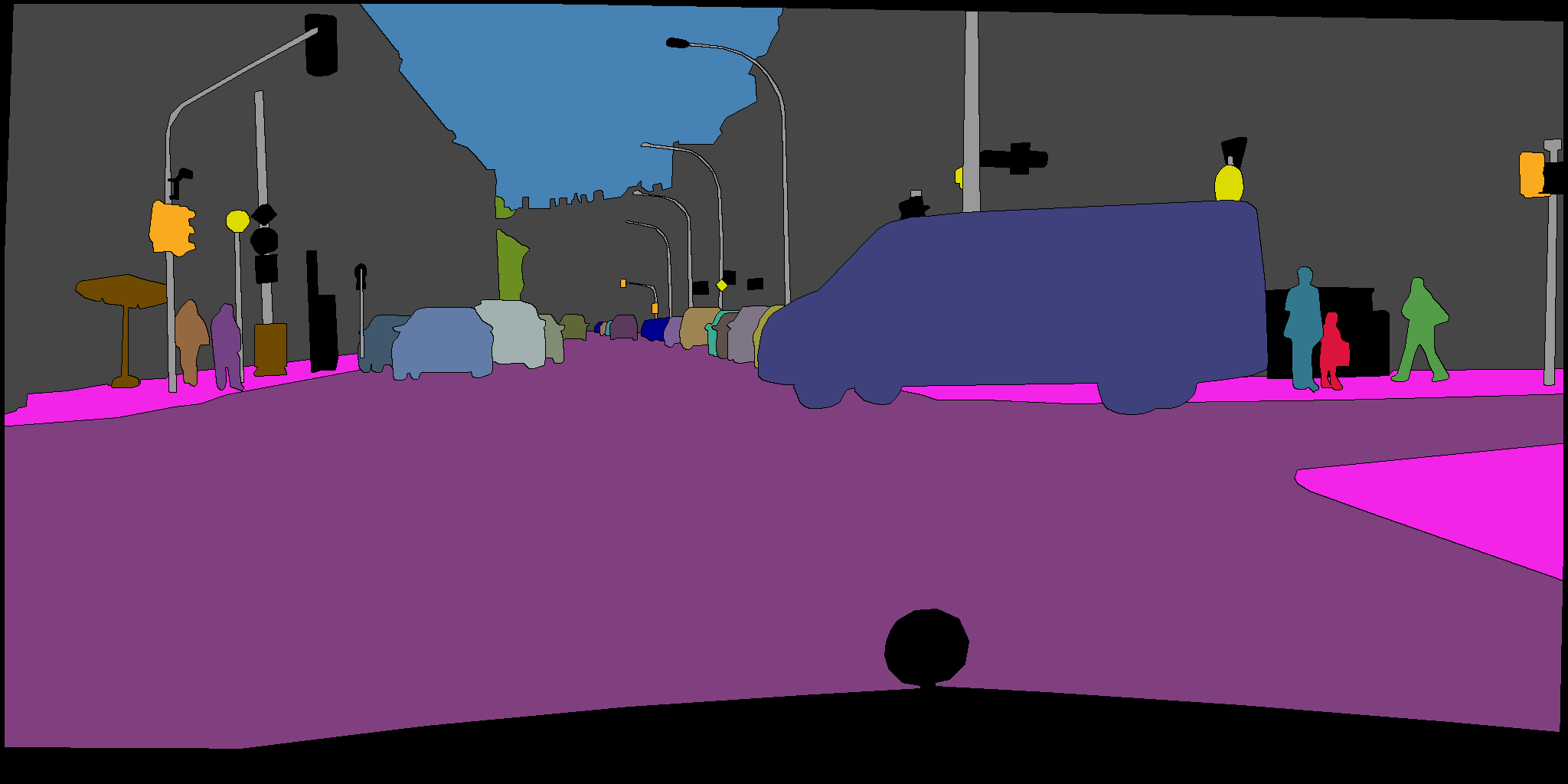} \hspace{4pt} &
\includegraphics[width=0.24\textwidth]{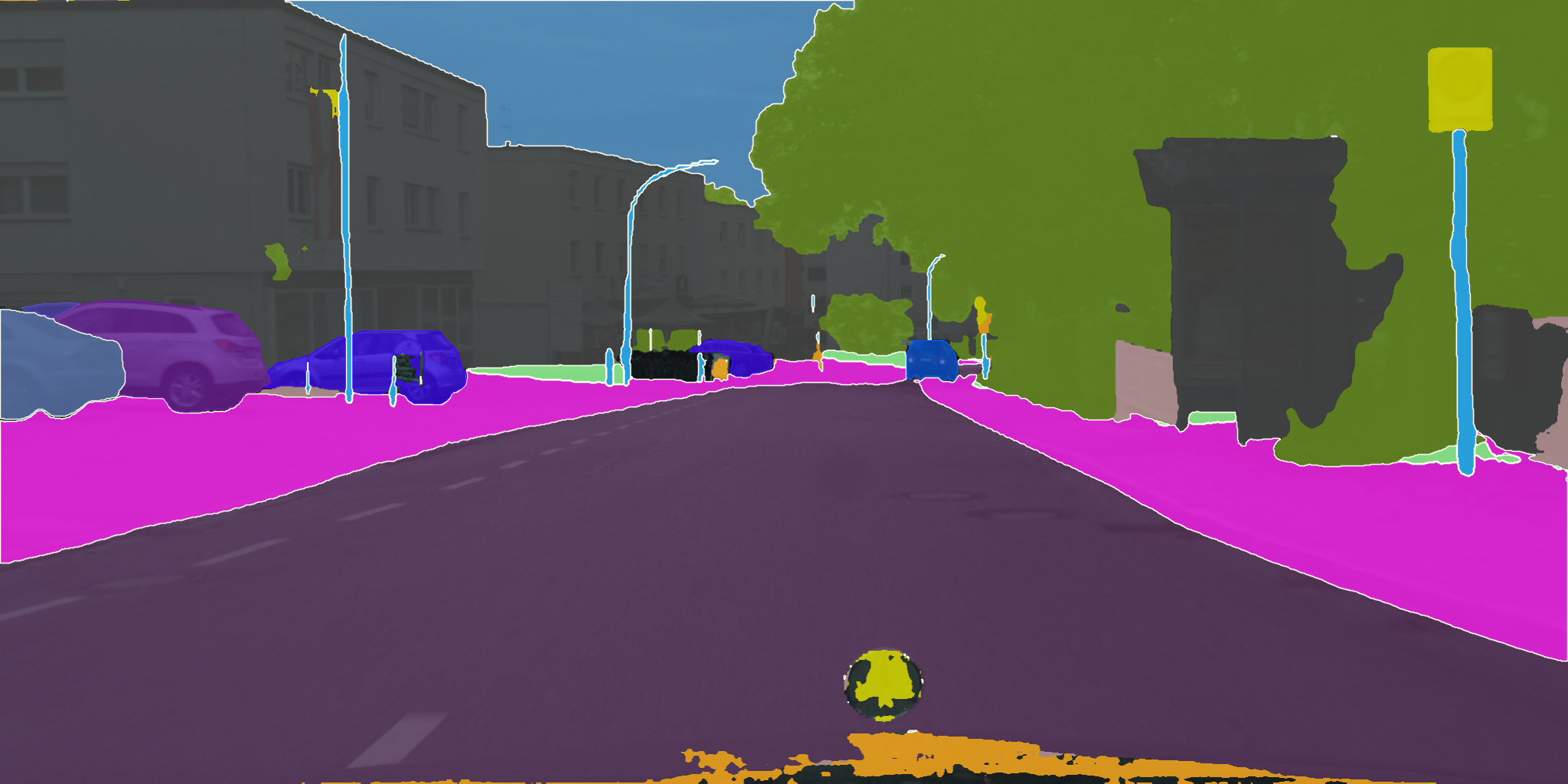} & 
\includegraphics[width=0.24\textwidth]{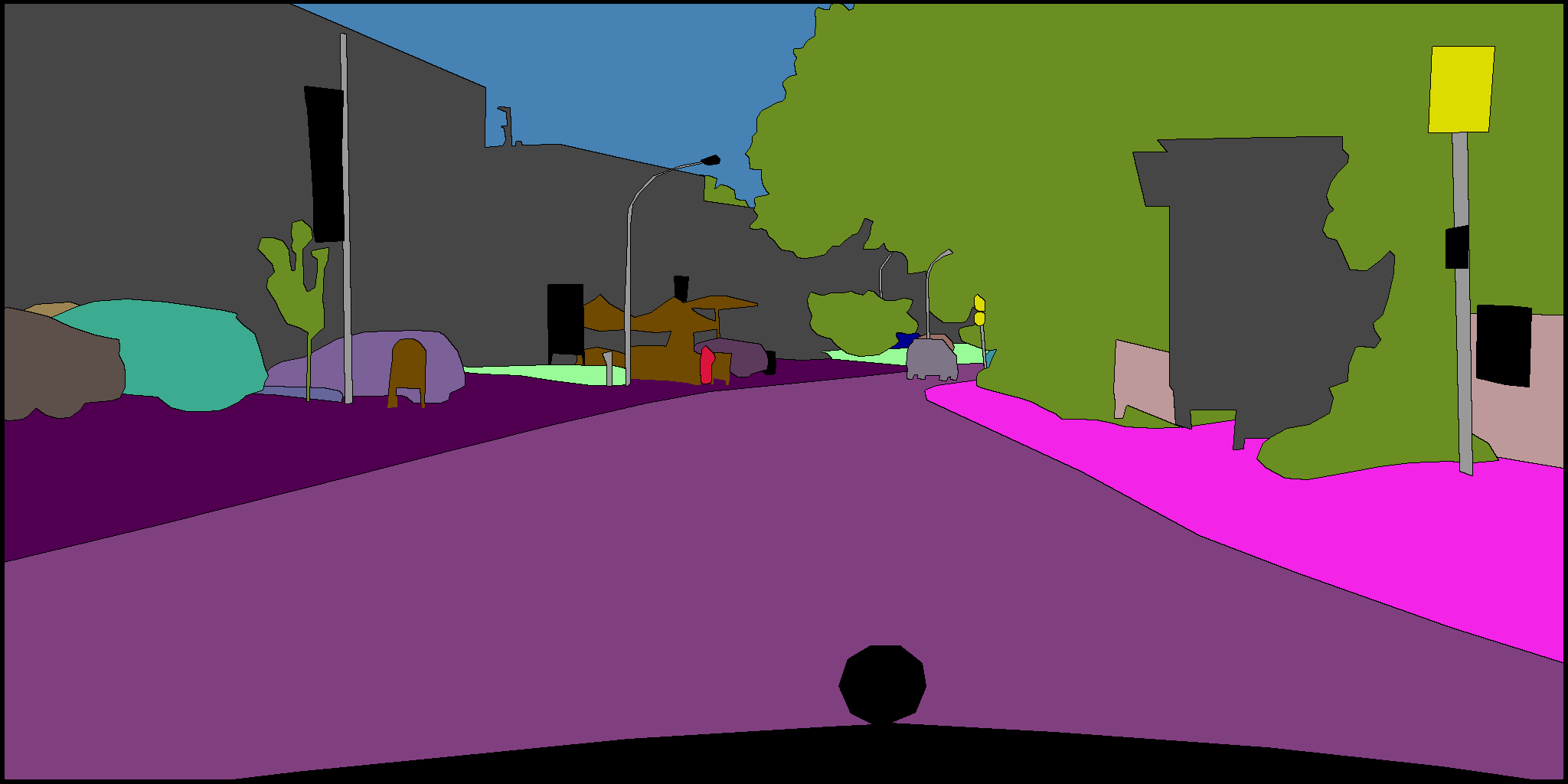} \vspace{4pt} \\

\includegraphics[width=0.24\textwidth]{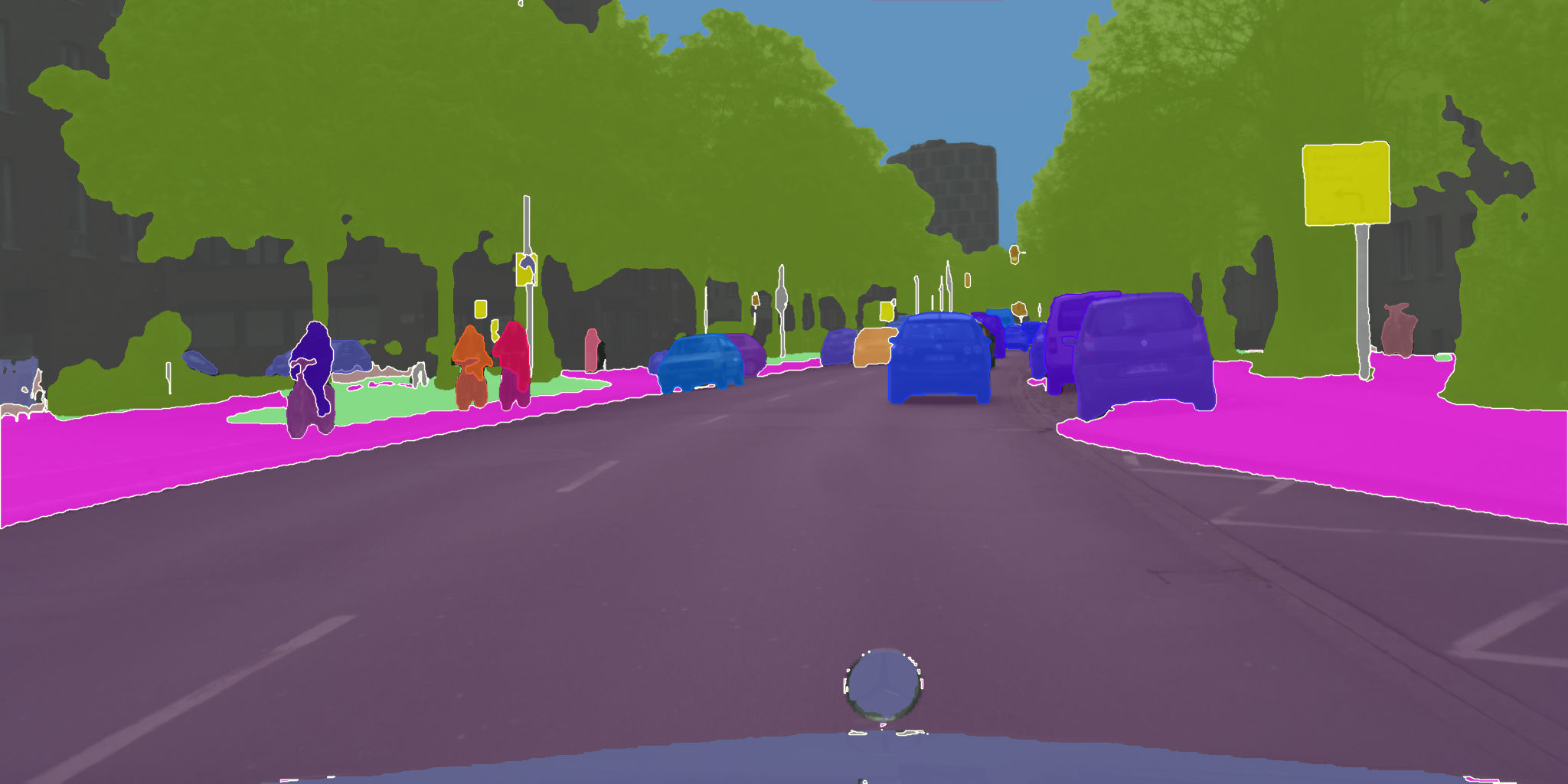} & 
\includegraphics[width=0.24\textwidth]{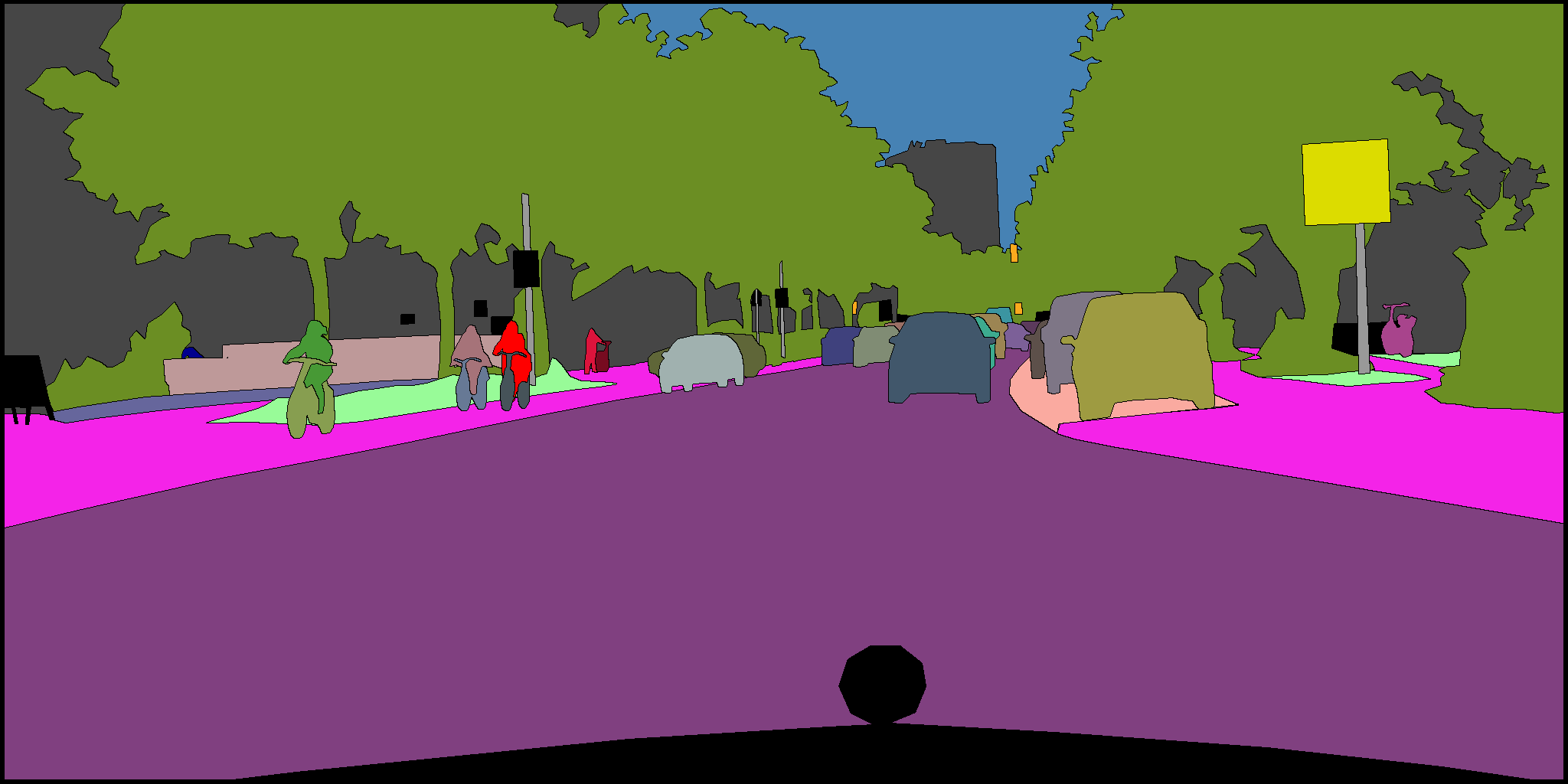} \hspace{4pt} &
\includegraphics[width=0.24\textwidth]{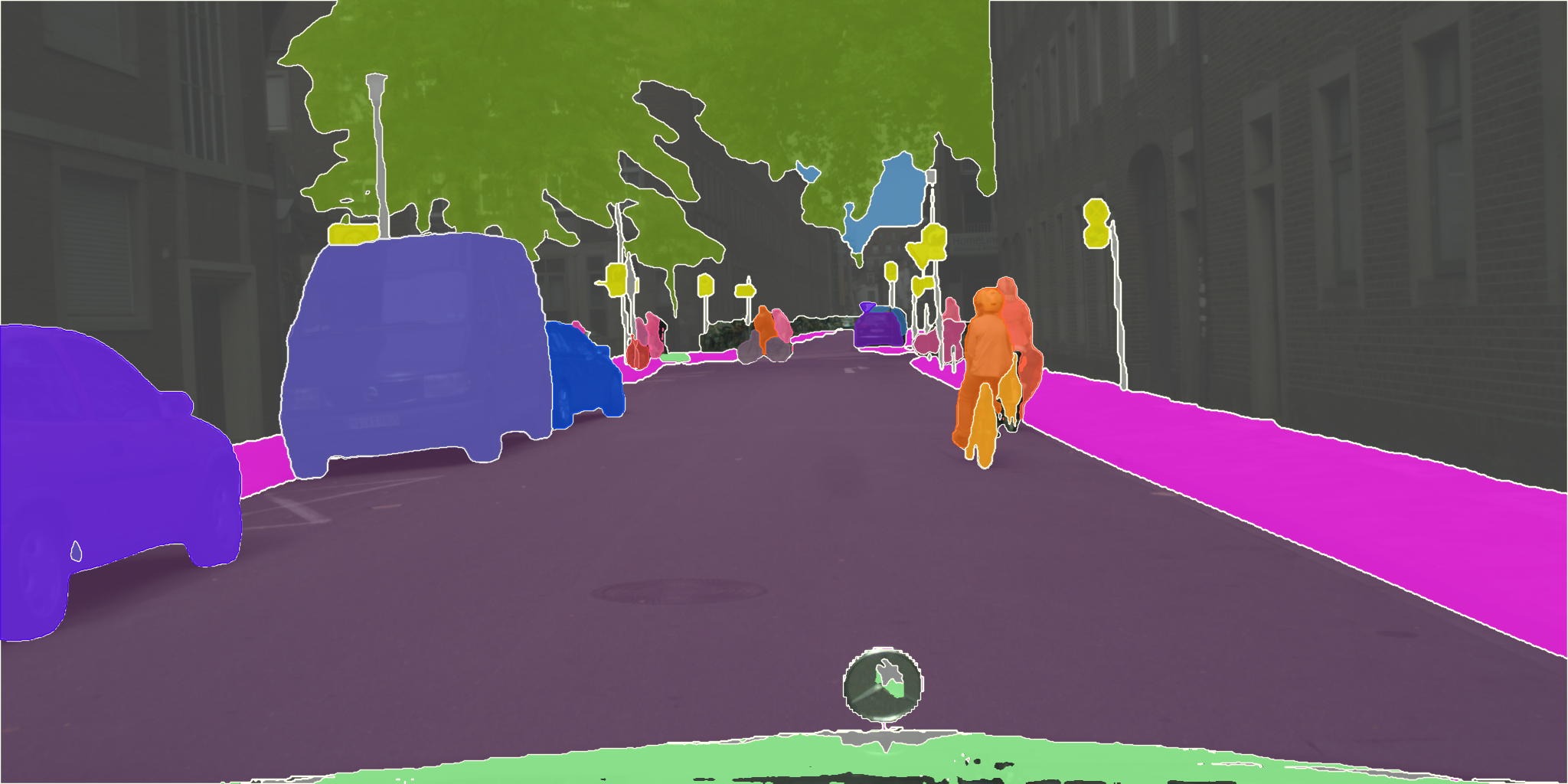} & 
\includegraphics[width=0.24\textwidth]{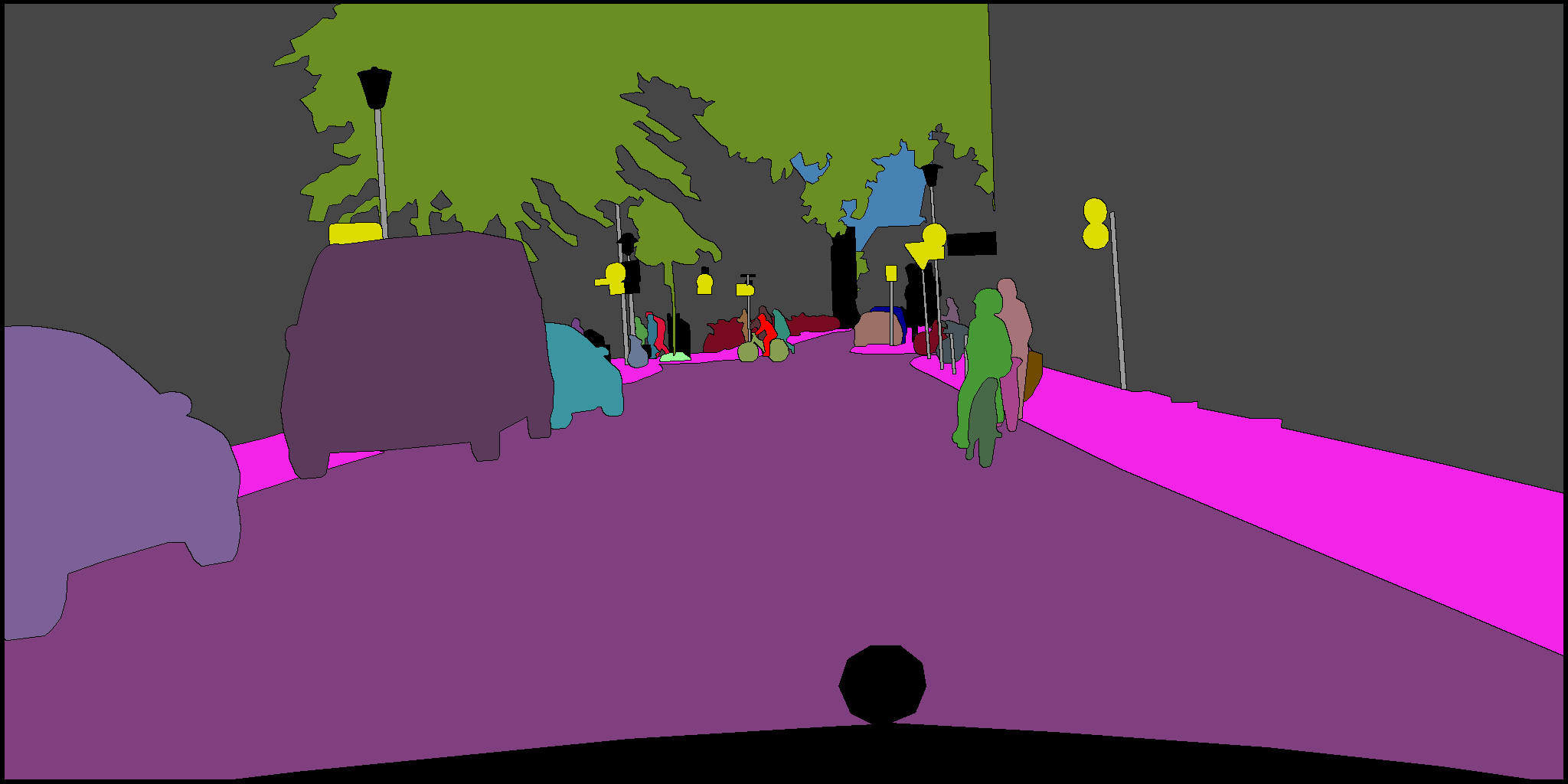}
\end{tabular}
\textbf{(a)} \\[4pt]

\begin{tabular}{@{}cccc@{}}
\includegraphics[width=0.24\textwidth]{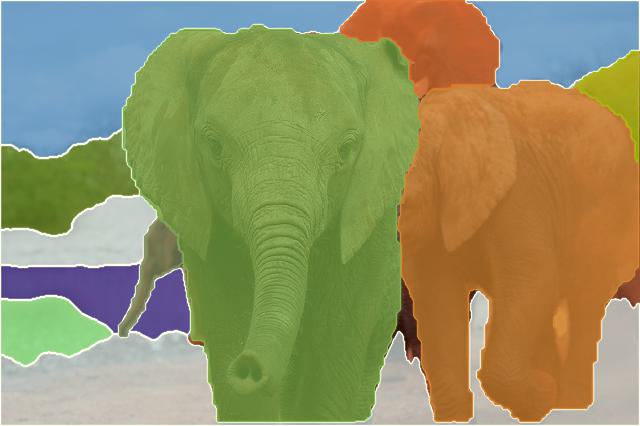} &
\includegraphics[width=0.24\textwidth]{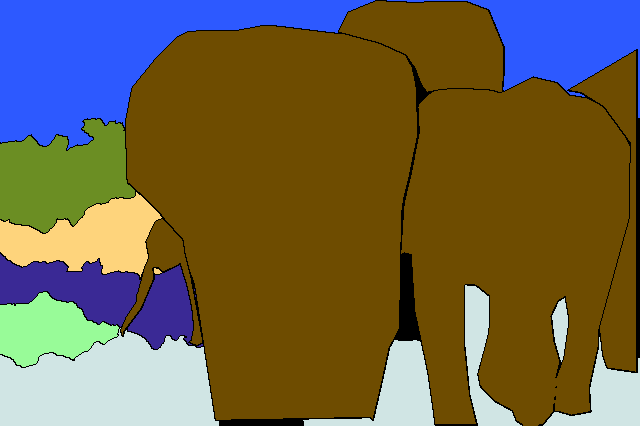} \hspace{4pt} &
\includegraphics[width=0.24\textwidth]{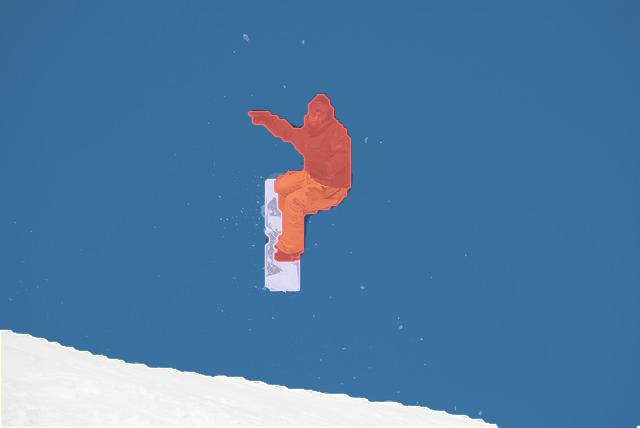} &
\includegraphics[width=0.24\textwidth]{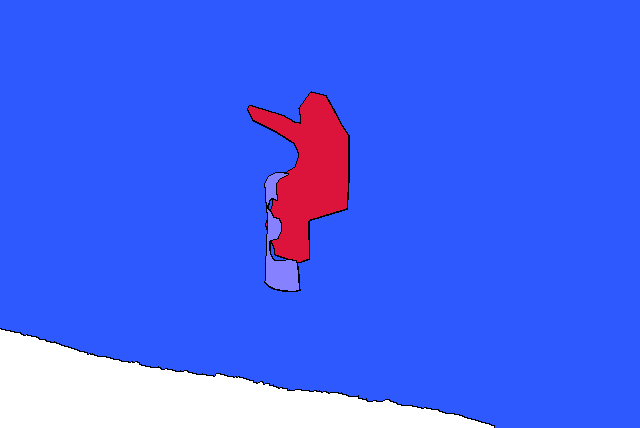} \vspace{4pt} \\

\includegraphics[width=0.24\textwidth]{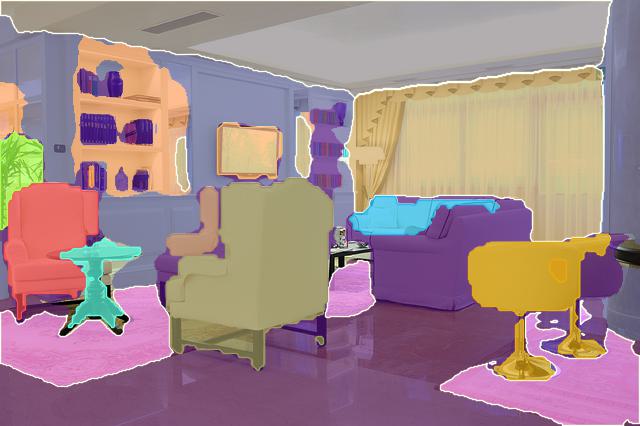} &
\includegraphics[width=0.24\textwidth]{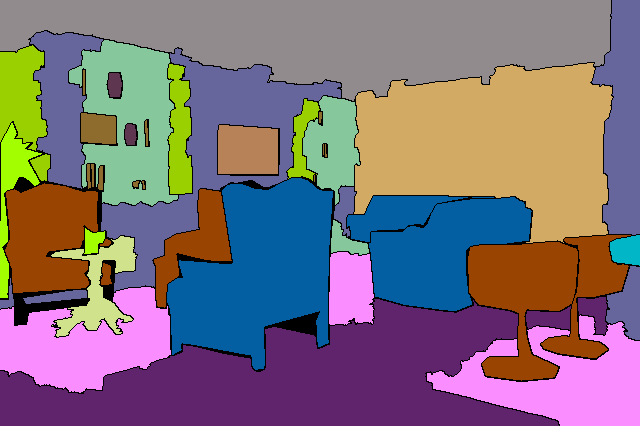} \hspace{4pt} &
\includegraphics[width=0.24\textwidth]{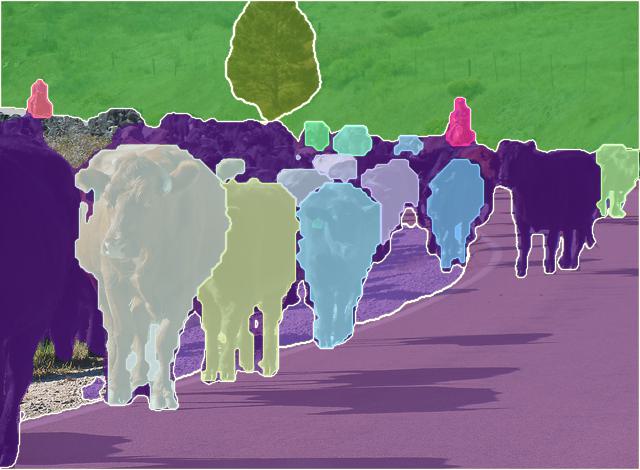} &
\includegraphics[width=0.24\textwidth]{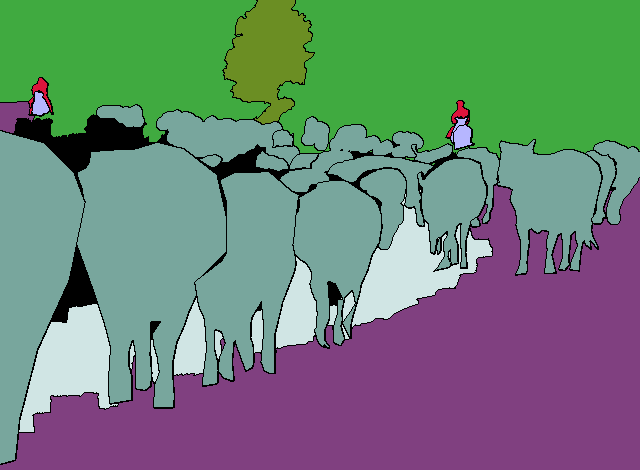}
\end{tabular}
\textbf{(b)} \\[4pt]
\caption{Examples of panoptic predictions on (a) Cityscapes validation set and (b) COCO validation. Each row has two examples. For each example, the prediction overlaid on the image is on the left and the ground truth is on the right.}
\label{fig:combined_panoptic_predictions}
\end{figure*}

\begin{figure*}[htbp]
\centering
\includegraphics[width=1.0\textwidth]{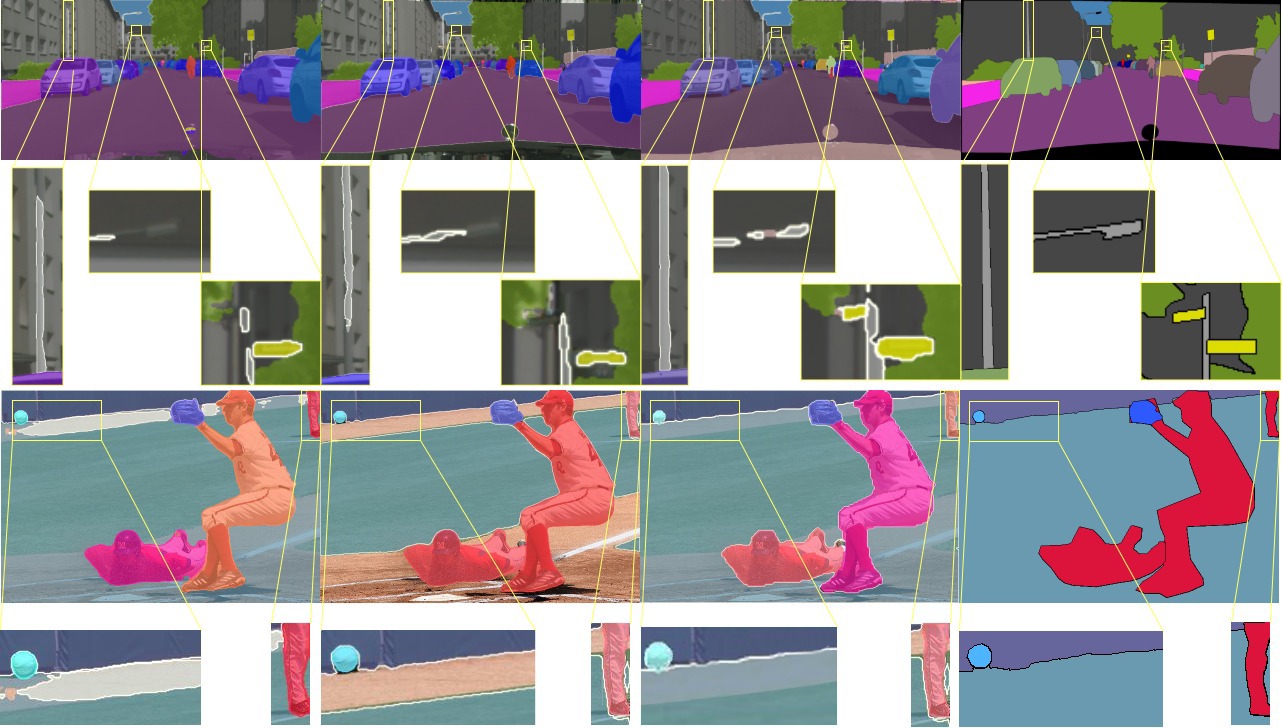}
\caption{Comparison of CNN-, transformer- and Mamba-based architectures. From left to right: Panoptic-DeepLab (ResNet-50), Mask2Former(Swin-T), our model and finally the ground truth. The two examples are from Cityscapes and COCO, respectively.}
\label{fig:comparison}
\end{figure*}

\section{EXPERIMENTAL RESULTS}
In this section, we present the experiments on two datasets: Cityscapes \cite{Cordts2016CityscapesCVPR} and COCO 2017 \cite{Lin2014COCO} for panoptic segmentation quantitatively and qualitatively. We compare against two CNN-based methods, PanopticFCN \cite{Li2021PanopticFCN} and Panoptic-DeepLab \cite{Cheng2020PanopticDeepLab}, and a transformer-based Mask2Former \cite{Cheng2022Mask2Former}. We further conduct ablation studies to assess the contribution of the proposed Mamba-based components. 

\subsection{Datasets}
Experiments are conducted on two widely adopted benchmarks for panoptic segmentation, specifically Cityscapes \cite{Cordts2016CityscapesCVPR} and COCO 2017 \cite{Lin2014COCO}.

The Cityscapes dataset is designed for urban scene understanding in the context of autonomous driving. Images were collected across 50 cities and are provided at a resolution of 2048×1024 pixels. The dataset comprises 5,000 finely annotated images with pixel-level panoptic labels: 2,975 for training, 500 for validation, and 1,525 for testing. It covers 19 semantic categories, of which 8 are \textit{thing} classes with unique instance labels and 11 are \textit{stuff} classes.

The COCO 2017 dataset (Lin et al., 2014b) is a large-scale benchmark for object detection, instance segmentation, and panoptic segmentation. It contains 118,000 training images, 5,000 validation images, and 20,000 test images, with dense panoptic annotations spanning 133 semantic categories: 80 \textit{thing} classes and 53 \textit{stuff} classes. 

\subsection{Implementation}
We implement our method with Detectron2 \cite{wu2019detectron2}. The base learning rate is set to 0.01, with gradient clipping applied to prevent gradient explosion. Network optimisation is performed using SGD with a weight decay of 1e-4, momentum of 0.9, and a batch size of 24, for a total of 90,000 iterations. The model is evaluated on the validation set every 5,000 iterations, and the checkpoint achieving the highest Panoptic Quality is selected for final evaluation, effectively serving as an early stopping criterion. Random cropping to 512×1024 pixels and CUDA Automatic Mixed Precision (AMP) are applied during training to reduce memory requirements.

\subsection{Metrics}
To evaluate the performance of panoptic segmentation, our main metric is the Panoptic Quality (PQ) \cite{Kirillov2019Panoptic}, a standard metric for the panoptic segmentation task, which jointly captures detection and segmentation performance. It is defined per image as: 

\vspace{-1em}
\begin{equation}
    \text{PQ} = \frac{\sum_{(p,g) \in TP} IoU(p, g)}{|TP| + \tfrac{1}{2}|FP| + \tfrac{1}{2}|FN|}
\end{equation}

where $TP$ is the set of true positive matches between predicted segments $p$ and ground-truth segments $g$ with $IoU > 0.5$, $FP$ is the false positives (predicted segments not matched to any ground truth), $FN$ is the false negatives (ground truth segments not matched to any prediction) and $IoU$ is the Intersection over Union between the matched prediction and ground truth.

Furthermore, we also report mean Intersection over Union (mIoU) for semantic segmentation and Average Precision (AP) for instance segmentation. These are defined as follows: 

\vspace{-1em}
\begin{equation}
    \mathrm{mIoU} = \frac{1}{N} \sum_{i=1}^{N} \frac{TP_i}{TP_i + FP_i + FN_i}
\end{equation}
where $TP_i$, $FP_i$, and $FN_i$ are respectively the true positives, false positives, and false negatives for class $i$, and $N$ is the number of classes.

\vspace{-1em}
\begin{equation}
    \mathrm{AP} = \int_{0}^{1} p(r)\, dr
\end{equation}
Average Precision (AP) is computed as the area under the prec\-ision-recall curve, where $p(r)$ is the precision at recall $r$.

\subsection{Results}

\begin{table}[h]
\centering
\caption{Evaluation results of Panoptic-DeepLab, PanopticFCN, Mask2Former and ours on Cityscapes $val$ set}
\resizebox{\columnwidth}{!}{%
\begin{tabular}{lc|ccc|cc|c}
\toprule
Method & Backbone & PQ & PQ$^{th}$ & PQ$^{st}$ & AP & mIoU & Param\\
\midrule
Panoptic-DeepLab & ResNet50 & 60.38 & 50.95 & 67.24 & 31.44 & 77.49 & 30.3M \\
PanopticFCN & ResNet50 & 59.6 & 52.1 & 65.1 & 32.2 & 76.8 & 36.6M \\
Mask2Former & Swin-T & 63.9 & 56.2 & 67.8 & 39.1 & \textbf{80.5} & 47.4M \\
Ours & SegMan & \textbf{64.89} & \textbf{58.26} & \textbf{69.70} & \textbf{39.91} & 80.18 & 35.7M\\
\bottomrule
\end{tabular}
}
\label{table:cityscapes}
\end{table}

\begin{table}[h]
\centering
\caption{Evaluation results of Panoptic-DeepLab, PanopticFCN, Mask2Former and ours on COCO $val$ set}
\resizebox{\columnwidth}{!}{%
\begin{tabular}{lc|ccc|cc|c}
\toprule
Method & Backbone & PQ & SQ & RQ & AP & mIoU & Param\\
\midrule
Panoptic-DeepLab & ResNet50 & 35.5 & 77.3 & 44.7 & 19.7 & 40.1 & 30.3M \\
PanopticFCN & ResNet50 & 41.0 & 81.0 & 49.6 & 30.7 & 43.6 & 36.6M \\
Mask2Former & Swin-T & \textbf{53.2} & \textbf{82.2} & \textbf{64.3} & \textbf{43.3} & \textbf{63.2} & 47.4M \\
Ours & SegMan & 45.6 & 81.5 & 54.7 & 34.2 & 46.4 & 35.7M\\
\bottomrule
\end{tabular}
}
\label{table:coco}
\end{table}

To ensure a fair comparison, all baselines and the proposed model are trained under identical conditions, including the same training schedules, data augmentation strategies, and input resolutions. Backbone networks are selected to maintain comparable parameter counts without modifying the original baseline architectures.

In Table \ref{table:cityscapes}, we present the quantitative evaluation results on the Cityscapes validation set. MambaPanoptic outperforms both CNN-based baselines across all reported metrics, and achieves higher PQ and AP than Mask2Former with a slightly lower mIoU, while using 11.7M fewer parameters than the transformer-based model. These results confirm that Mamba-based architectures offer a strong and efficient alternative to both convolutional and transformer designs for panoptic segmentation on urban scene benchmarks.

In Table \ref{table:coco}, we present the quantitative evaluation results on the COCO validation set. MambaPanoptic consistently outperforms both CNN-based baselines but shows a performance gap relative to Mask2Former.  This gap can be attributed in part to the substantially greater category complexity of COCO. COCO has 80 \textit{thing} classes and 53 \textit{stuff} classes while Cityscapes only has 8 \textit{thing} classes and 11 \textit{stuff} classes. Mask2Former \cite{Cheng2022Mask2Former} employs 100 learnable queries for COCO panoptic segmentation, each of which can specialise to a distinct category after training, affording high representational capacity for large-vocabulary scenes. In contrast, both MambaPanoptic and PanopticFCN \cite{Li2021PanopticFCN} generate kernels dynamically from each input image, without the benefit of persistent, category-specific query representations. This architectural distinction means that query-based methods carry a representational advantage in high-category settings, at the cost of heavier decoders and quadratic self-attention.

In Figure \ref{fig:combined_panoptic_predictions}, we present qualitative panoptic segmentation results on both validation sets.
The proposed model correctly segments both \textit{thing} instances and \textit{stuff} regions, including objects that are distant or partially occluded. Predictions near object boundaries occasionally exhibit imprecision. Figure \ref{fig:comparison} provides a direct qualitative comparison of small and distant structures in both datasets among Panoptic-DeepLab, Mask2Former, and MambaPanoptic, which shows the effectiveness of our proposed method compared to Panoptic-DeepLab and Mask2Former.

\subsection{Ablations}

\begin{table}[h]
\centering
\caption{Ablation results of Mamba-based modules on the Cityscapes $val$ set.}
\resizebox{\columnwidth}{!}{%
\begin{tabular}{l|ccc|cc}
\toprule
Method & PQ & PQ$^{th}$ & PQ$^{st}$ & AP & mIoU \\
\midrule
w/o Mamba Encoder  & 59.85 & 52.43 & 64.49 & 32.54 & 77.01 \\
w/o MambaFPN & 62.49 & 53.42 & 66.09 & 34.82 & 78.14 \\
w/o QuadMamba & 64.39 & 58.03 & 69.28 & 39.13 & 79.95 \\
Ours  & \textbf{64.89} & \textbf{58.26} & \textbf{69.70} & \textbf{39.91} & \textbf{80.18} \\
\bottomrule
\end{tabular}
}

\label{table:ablation}
\end{table}

In this ablation section, we mainly verify the effectiveness of the three Mamba-based modules introduced. For each ablation, we use the same training and evaluation settings except for the replacement of the test module. In Table \ref{table:ablation}, we show the ablation results on the Cityscapes dataset. \textit{Ours} represents the proposed model; \textit{w/o Mamba Encoder} represents the model that uses the ResNet50 as the backbone instead of the SegMan encoder; \textit{w/o MambaFPN} represents the model that uses CNN FPN instead of the MambaFPN for multi-scale feature generation; \textit{w/o QuadMamba} represents the model that uses a CNN instead of the QuadMamba for feature refinement. The results in Table \ref{table:ablation} confirm that each proposed Mamba-based module contributes positively and independently to the overall panoptic segmentation performance.

\section{CONCLUSION}

In this paper, we have presented MambaPanoptic, a Mamba-based framework for panoptic segmentation. The proposed architecture combines a multi-scale feature encoder based on the SegMAN backbone and the proposed MambaFPN, together with a PanopticFCN-style kernel-based panoptic head. To further enhance feature quality, we incorporated a QuadMamba-based refinement module into both kernel generation and high-resolution feature refinement. Experimental results on the Cityscapes and COCO panoptic benchmarks demonstrate that the proposed method is effective and competitive, showing clear improvements over strong CNN-based baselines and strong performance on Cityscapes with fewer parameters than Mask2Former. These results suggest that state-space models are a promising alternative to conventional convolutional and transformer-based architectures for panoptic segmentation.

Despite these encouraging results, the proposed method still has several limitations. First, as observed in the qualitative results, the model occasionally struggles near object boundaries and thin structures. We hypothesize that this behaviour is attributable to a trade-off inherent in Mamba-based scanning mechanisms, which are effective at modeling long-range dependencies but may be less suited to capturing high-frequency local details. Incorporating boundary-aware supervision or more specialized high-resolution refinement may help alleviate this issue. Second, although the kernel-based head is efficient, the method still lags behind stronger query-based transformer models on highly complex datasets such as COCO, which contain many semantic categories and crowded scene layouts. This suggests that query-based decoding may provide higher representation capacity in such settings. Exploring a hybrid design that combines efficient Mamba-based feature modeling with a lightweight query-based decoder is therefore a promising direction for future work.

In future work, we plan to investigate larger Mamba-based backbones, improve prediction quality near boundaries and thin structures, explore more tightly integrated Mamba-based pixel decoders and panoptic heads, and extend the framework to related dense prediction tasks such as aerial-image and 3D panoptic segmentation, where efficient long-range context modeling may offer even greater benefits.

{
\begin{spacing}{1.17}
    \normalsize
    \bibliography{references}
\end{spacing}
}

\end{document}